\documentclass[11pt, a4paper, logo, copyright, nonumbering]{wechat}
\usepackage[numbers, square, sort&compress]{natbib}
\usepackage{dblfloatfix}
\usepackage{ulem}
\usepackage{caption}
\definecolor{citecolor}{HTML}{1976D2}
\hypersetup{
    colorlinks=true,
    linkcolor=black,
    filecolor=magenta,
    urlcolor=blue!50!black,
    citecolor=citecolor,
}
\usepackage{dramatist}
\usepackage{xspace}
\usepackage{pifont}% http://ctan.org/pkg/pifont
\usepackage{multirow}
\usepackage{tcolorbox}
\usepackage{xltabular}
\usepackage{longtable}
\usepackage{hyperref}
\usepackage{diagbox}
\usepackage{makecell}
\usepackage{amsfonts}
\usepackage{amsmath}
\usepackage{amssymb}
\usepackage{lineno}

\usepackage[bottom]{footmisc}
\usepackage{CJKutf8}
\usepackage{subfigure}
\usepackage{setspace}
\usepackage{subcaption} 

\usepackage{titlesec}

\usepackage{tcolorbox}
\usepackage{fontawesome5}
\usepackage{enumitem}

\usepackage{booktabs}
\usepackage{multirow}
\usepackage[table,xcdraw]{xcolor}
\usepackage{graphicx}

\usepackage{algorithm}
\usepackage{algpseudocode}

\tcbset{
    contributionbox/.style={
        colback=blue!3,
        colframe=blue!50!black,
        boxrule=0.5pt,
        arc=3pt,
        left=6pt,
        right=6pt,
        top=6pt,
        bottom=6pt,
        fonttitle=\bfseries
    }
}

\titlespacing*{\paragraph}
  {0pt}                   
  {0.5ex plus 1ex minus .2ex} 
  {1em}            

\makeatletter
\def\@BTrule[#1]{%
  \ifx\longtable\undefined
    \let\@BTswitch\@BTnormal
  \else\ifx\hline\LT@hline
    \nobreak
    \let\@BTswitch\@BLTrule
  \else
     \let\@BTswitch\@BTnormal
  \fi\fi
  \global\@thisrulewidth=#1\relax
  \ifnum\@thisruleclass=\tw@\vskip\@aboverulesep\else
  \ifnum\@lastruleclass=\z@\vskip\@aboverulesep\else
  \ifnum\@lastruleclass=\@ne\vskip\doublerulesep\fi\fi\fi
  \@BTswitch}
\makeatother

\addto\extrasenglish{
}

 {\begin{list}{}%
         {\setlength{\leftmargin}{#1}}%
         \item[]%
 }
 {\end{list}}

\reportnumber{001} % Leave blank if n/a

\renewcommand{\today}{}

\title{\centering WeDLM: Reconciling Diffusion Language Models with Standard Causal Attention for Fast Inference}

\author{
    \vspace{0.5em}
    {\large \bfseries
    Aiwei Liu$^{1, \ast, \dagger}$, Minghua He$^{1, 2, \ast, \ddagger}$, Shaoxun Zeng$^{3}$, Sijun Zhang $^{1}$, 
    } \\ \vspace{5pt}
    {\large \bfseries
    Linhao Zhang$^{1}$, Chuhan Wu$^{1}$, Wei Jia$^{1}$, Yuan Liu$^{1}$, Xiao Zhou$^{1}$, Jie Zhou$^{1}$
    } \\ \vspace{1.0em}

    {\normalsize \normalfont
    $^{1}$WeChat AI, Tencent \quad $^{2}$Peking University \quad $^{3}$Tsinghua University
    }
}

\renewcommand{\phi}{\varphi}

\renewcommand{\leq}{\leqslant}

\renewcommand{\epsilon}{\varepsilon}
\renewcommand{\imath}{\mathrm{i}}

\newlength{\restsubwidth}
\newlength{\restsubheight}
\newlength{\restsubmoreheight}
\newcommand{\rest}[2]{%
        \settowidth{\restsubwidth}{\ensuremath{#2}}
        \settoheight{\restsubheight}{\ensuremath{{}_{#2}}}
        \ensuremath{{#1\hskip 0.5pt}_{\vrule\kern2pt\parbox[b][%
        4pt][b]{\the\restsubwidth}{%
                        \ensuremath{{}_{#2}}}}}
        }

\projectpage{https://wedlm.github.io}
\github{https://github.com/tencent/WeDLM}
\huggingface{https://huggingface.co/collections/tencent/wedlm}

\begin{abstract}
Autoregressive (AR) generation is the standard decoding paradigm for Large Language Models (LLMs), but its token-by-token nature limits parallelism at inference time. Diffusion Language Models (DLLMs) offer parallel decoding by recovering multiple masked tokens per step; however, in practice they often fail to translate this parallelism into deployment speed gains over optimized AR engines (e.g., vLLM). A key reason is that many DLLMs rely on bidirectional attention, which breaks standard prefix KV caching and forces repeated contextualization, undermining efficiency.
We propose \textbf{\texttt{WeDLM}}, a diffusion decoding framework built entirely on \emph{standard causal attention} to make parallel generation prefix-cache friendly. The core idea is to let each masked position condition on all currently observed tokens \emph{while keeping a strict causal mask}, achieved by \textit{Topological Reordering} that moves observed tokens to the physical prefix while preserving their logical positions. Building on this property, we introduce a streaming decoding procedure that continuously commits confident tokens into a growing left-to-right prefix and maintains a fixed parallel workload, avoiding the stop-and-wait behavior common in \emph{block diffusion} methods. Experiments show that \texttt{WeDLM} preserves the quality of strong AR backbones while delivering substantial speedups, approaching $3\times$ on challenging reasoning benchmarks and up to $10\times$ in low-entropy generation regimes; critically, \textbf{our comparisons are against AR baselines served by vLLM under matched deployment settings}, demonstrating that diffusion-style decoding can outperform an optimized AR engine in practice.
\end{abstract}

\begin{document}
\begin{CJK*}{UTF8}{gbsn}

\maketitle

\enlargethispage{1cm}

\newcommand\blfootnote[1]{%
  \begingroup
  \renewcommand\thefootnote{}\footnote{#1}%
  \addtocounter{footnote}{-1}%
  \endgroup
}

\blfootnote{$^\ast$ Equal Contribution.}
\blfootnote{$^\dagger$ Corresponding to Aiwei Liu: \texttt{coveliu@tencent.com}}
\blfootnote{$^\ddagger$ Work done during internship at WeChat AI.}

\vspace{-2em}

\begin{figure}[h]
    \centering
    \includegraphics[width=0.98\textwidth]{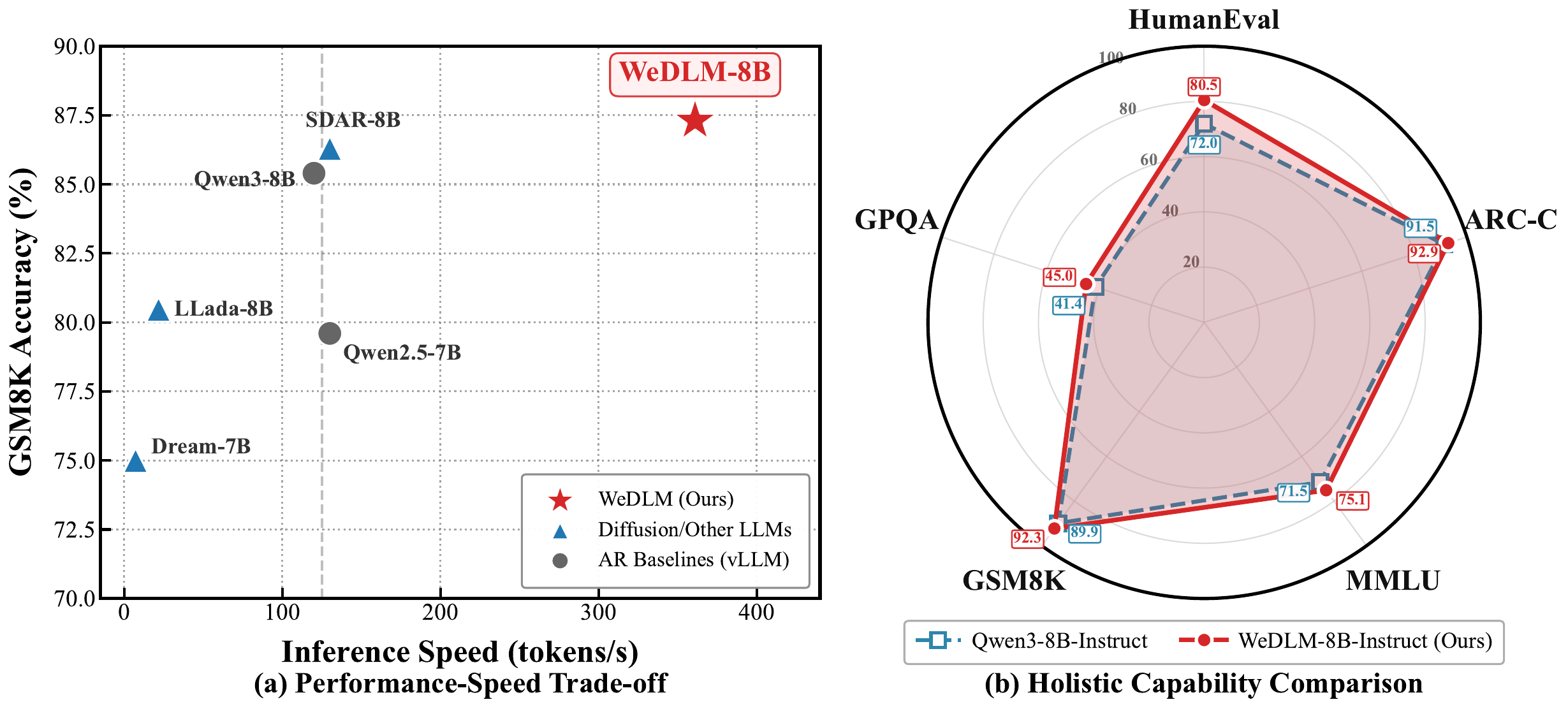}

\caption{
\textbf{Performance and capability overview of \texttt{WeDLM-8B}.}
(a) \textbf{Speed vs.\ Accuracy:} \texttt{WeDLM-8B} achieves a $\sim$3$\times$ speedup over the vLLM-optimized AR baseline (\texttt{Qwen3-8B}) on GSM8K, while also significantly outperforming prior diffusion models in both inference speed (tps) and accuracy. Dream and LLaDA use the dInfer inference engine \citep{ma2025dinfer}; SDAR uses JetEngine.
(b) \textbf{Holistic Evaluation:} \texttt{WeDLM-8B-Instruct} matches or surpasses the strong capabilities of the \texttt{Qwen3-8B-Instruct} baseline, showing improvements across several mathematical, coding, and general knowledge benchmarks.
}
    \label{fig:1}
\end{figure}

\section{Introduction}

The autoregressive (AR) generation of Large Language Models (LLMs) is bottlenecked by its step-by-step nature. This sequential decoding underutilizes modern accelerators and often becomes memory bound \citep{dao2022flashattention}. Diffusion Language Models (DLLMs) offer an appealing alternative: they recover multiple masked tokens in parallel \citep{zhang2025survey}. Yet in practical deployments, existing DLLMs have not shown a clear speed advantage over highly optimized AR serving engines such as vLLM \citep{vllm}. A key reason is that AR systems convert algorithmic efficiency into real throughput via \emph{native KV caching} and mature runtime optimizations (e.g., PagedAttention \citep{vllm} and CUDA Graphs). This implies that outperforming optimized AR baselines requires more than parallel prediction: a DLLM must be \emph{prefix-cache compatible}, i.e., it should continuously grow a cache-valid left-to-right prefix so that most computation is reused rather than recomputed.

The main obstacle is prefix-cache incompatibility in prior diffusion designs. Many representative DLLMs (e.g., LLaDA~\citep{nie2025large} and Dream~\citep{ye2025dream}) employ full bidirectional attention, which intrinsically couples each token’s representation to both past and future positions. As a result, KV caching cannot be applied in the standard way: even early predictions are not immediately cache-valid and must be recomputed as later tokens change. Block-wise variants (e.g., SDAR~\citep{cheng2025sdar} and NBDiff~\citep{tian2025next}) partially restore prefix reuse by committing completed blocks, but their gains are constrained by two effects. First, bidirectional attention \emph{within} a block delays cacheability: a token cannot be committed until the entire block is finalized, because its KV state may depend on unresolved suffix positions. Second, diffusion-style resolution can be out of order, which reduces how much of the newly predicted content forms a contiguous left-to-right prefix and therefore limits effective cache reuse. These observations motivate a different design choice: we argue that bidirectional attention is not essential for parallel mask recovery, and that restoring strict causal structure is the most direct path to cache-friendly diffusion decoding.

In this work, we propose \textbf{\texttt{WeDLM}}, a framework that performs diffusion-style mask recovery entirely under \emph{standard causal attention} to make parallel decoding compatible with prefix caching. Our key insight is that mask recovery only requires each masked position to access all currently observed tokens; this can be achieved without bidirectional attention via \textit{Topological Reordering}. Specifically, we move observed tokens to the physical front while preserving their logical positions through RoPE position ids~\citep{su2024roformer}, so masked tokens can attend to the full observed context under an unmodified causal mask. This causal structure is naturally aligned with prefix caching: once earlier positions are resolved, their KV states depend only on committed context and can be reused immediately. We further introduce \textit{Dual-Stream Masking} to reduce the training--inference gap induced by prefix-conditioned decoding. By constructing a clean \textit{memory stream} alongside a masked \textit{prediction stream} (with shared positional encoding), each prediction block is trained to condition on clean history rather than on potentially noisy intermediate predictions.

For inference, we develop \textit{Streaming Parallel Decoding}, an algorithm explicitly organized around \emph{prefix commitment}. It combines: (i) a position-aware confidence rule (implemented as a distance-penalized selection) that prioritizes earlier unresolved positions and encourages left-to-right growth; (ii) strict causal attention, which guarantees that newly committed prefix tokens become cache-valid immediately; and (iii) a dynamic sliding window that continuously refills new masked slots as soon as tokens are committed, avoiding the stop-and-wait behavior of block-wise methods. With attention remaining a standard causal mask, each iteration reduces to a small causal prefill over the active window on top of an existing KV cache, enabling direct use of optimized AR inference infrastructure such as FlashAttention~\citep{dao2022flashattention}, PagedAttention~\citep{vllm}, and CUDA Graphs without kernel changes.

Experimental results demonstrate that \texttt{WeDLM} efficiently adapts to standard AR backbones. We instantiate \texttt{WeDLM} on both Qwen2.5-7B and Qwen3-8B, utilizing 100B tokens for continued training and 10B tokens for SFT. Across diverse benchmarks, including code generation (MBPP~\cite{austin2021program}, HumanEval~\cite{chen2021evaluating}, HumanEval-plus~\cite{liu2023code}), math reasoning (GSM8K~\cite{cobbe2021training}, MATH~\cite{hendrycks2020measuring}, GPQA~\cite{rein2024gpqa}), and general knowledge (MMLU~\cite{hendrycks2021mmlu}, ARC~\cite{clark2018think}, HellaSwag~\cite{zellers2019hellaswag}), \texttt{WeDLM} not only preserves but often improves upon the capabilities of its base models. Notably, \texttt{WeDLM-8B} achieves an average score of 77.36 on our benchmark suite, surpassing Qwen3-8B-Instruct (75.12) by over 2 points. \textbf{Unlike prior works that compare against unoptimized baselines, we benchmark \texttt{WeDLM} directly against the state-of-the-art vLLM engine.} Results show that \texttt{WeDLM} achieves up to 3$\times$ end-to-end acceleration on complex reasoning tasks and exceeds 10$\times$ speedups on some low-entropy generation scenarios, \textbf{demonstrating that diffusion-style decoding can outperform an optimized AR engine in matched, practical inference conditions}.

\begin{tcolorbox}[contributionbox, title={\faLightbulb~Main Contributions}]
\begin{itemize}[leftmargin=*, itemsep=6pt, label=\textcolor{blue!70!black}{\faCheckCircle}]
    \item \textbf{Causal Diffusion.} We propose \texttt{WeDLM}, a DLLM framework that performs mask recovery entirely under causal attention via \textit{Topological Reordering}. This design enables seamless initialization from pre-trained AR checkpoints and inherent prefix-cache compatibility---predicted tokens can be cached immediately without waiting for subsequent positions.
    \item \textbf{Streaming Parallel Decoding.} We introduce a decoding strategy designed around prefix-cache compatibility: a distance penalty promotes left-to-right resolution, the causal mask enables immediate caching of predicted prefixes, and a dynamic sliding window continuously refills new masks as finalized tokens are committed---eliminating the stop-and-wait bottleneck of block-wise methods.
    \item \textbf{First DLLM to Outperform Industrial AR Engines.} We demonstrate that \texttt{WeDLM} surpasses optimized vLLM baselines in wall-clock speed, achieving more than 3$\times$ speedups on complex reasoning tasks while maintaining generation quality.
\end{itemize}
\end{tcolorbox}

\section{Preliminary}
\label{sec:preliminary}

\subsection{Autoregressive Language Modeling}
\label{subsec:ar_lm}

Given a token sequence $\mathbf{x} = [x_1, x_2, \dots, x_T]$, a standard autoregressive language model factorizes the joint probability into conditional probabilities:
\begin{equation}
    P(\mathbf{x}) = \prod_{t=1}^{T} P(x_t \mid x_{<t}; \theta),
    \label{eq:ar_factorization}
\end{equation}
where $\theta$ denotes model parameters and $x_{<t} = [x_1, \dots, x_{t-1}]$ is the preceding context. During decoding, the model generates $x_t$ conditioned on the previously generated prefix. Consequently, generation is sequential: later tokens depend on earlier ones through the conditioning structure in Eq.~\ref{eq:ar_factorization}.

\subsection{Decoupled Positional Representation}
\label{subsec:decoupled_position}

Eq.~\ref{eq:ar_factorization} implicitly ties each token's \emph{logical position} to its \emph{physical index} (i.e., $p_t=t$). We make this dependency explicit by representing inputs as token--position pairs $(x_t, p_t)$. The model then defines the conditional distribution as a function of both token identities and supplied position ids:
\begin{equation}
    P(x_t \mid x_{<t}, p_{\leq t}; \theta) = \text{LLM}(x_{\le t}, p_{\le t}; \theta),
    \label{eq:decoupled_lm}
\end{equation}
where $\text{LLM}(\cdot)$ denotes the model-induced conditional distribution at position $t$ (implemented by a softmax over the logits at that position). This decoupling is naturally supported by Rotary Positional Embeddings (RoPE)~\cite{su2024roformer}, where attention scores are indexed by the supplied logical positions $\mathbf{p}$ rather than by physical indices. Therefore, tokens may be processed in a different physical order while still being referenced by their logical positions; the resulting computation additionally depends on the attention mask, which determines the allowed information flow. We exploit this flexibility in our method.

\subsection{Masked Diffusion Language Models}
\label{subsec:mdlm}

Masked Diffusion Language Models (MDLMs) formulate text generation as denoising rather than strictly sequential prediction. Given a clean sequence $\mathbf{x}_0$ of length $L$, a noising process samples a masking ratio $\gamma \in (0, 1]$ and corrupts a random subset of positions $\mathcal{M}$ (with $|\mathcal{M}| = \gamma L$) by replacing them with a special \texttt{[MASK]} token, producing a corrupted sequence $\mathbf{x}_\gamma$. The model is trained to reconstruct the original tokens at the masked positions.
A commonly used training objective is a weighted cross-entropy loss:
\begin{equation}
    \mathcal{L}(\theta) = -\mathbb{E}_{\gamma, \mathbf{x}_0, \mathbf{x}_\gamma} \left[ \frac{1}{\gamma} \sum_{i=1}^{L} \mathbf{1}[x_\gamma^{(i)} = \mathbf{M}] \log p_\theta(x_0^{(i)} \mid \mathbf{x}_\gamma) \right],
    \label{eq:mdlm_loss}
\end{equation}
where the factor $1/\gamma$ compensates for varying numbers of masked tokens under different noise levels. For brevity, we omit explicit conditioning on $\gamma$ (or the timestep) in $p_\theta$.

Standard MDLMs typically employ bidirectional attention so that masked positions can aggregate information from all observed tokens. However, this design introduces two limitations. First, bidirectional attention is incompatible with the KV cache mechanism that enables efficient autoregressive decoding. Second, when adapting pre-trained autoregressive models, the bidirectional structure induces an inductive-bias mismatch with causal representations. In this work, we optimize an MDLM-style denoising objective under strictly causal attention, enabled by reordering-based context exposure introduced in \S\ref{sec:method}.

\section{Motivation and Analysis}
\label{sec:analysis}
Two observations directly shape \texttt{WeDLM}'s design.  
(1) In KV-cached deployment, decoding speed is governed primarily by \emph{prefix cacheability} rather than per-step parallelism.  
(2) Mask recovery does not require bidirectional attention; it can be implemented with \emph{standard causal attention}.  

\subsection{Prefix Cacheability ($p_{\text{cache}}$) as an Inference Metric}
\label{subsec:prefix_cacheability}

Prior DLLMs mostly pursue speed by increasing \emph{tokens predicted per forward}. In practice, a more inference-critical factor is how many predicted tokens can be converted into a \emph{growing, KV-cache-valid prefix}. With KV caching, a token is reusable only if its key/value states depend \emph{only} on earlier context; therefore, \emph{only a left-to-right prefix is cacheable}. Predicted tokens that do not enter the committed prefix must be \emph{recomputed} in later forwards, increasing total compute.

We quantify this effect with two indicators. Let $N_{\text{gen}}$ be the number of \emph{new} tokens finally produced (excluding the initial prefill prompt), and let $N_{\text{fwd}}$ be the total number of token instances processed by the network across all decoding forwards after prefill (including repeated processing due to recomputation). We define the \emph{prefix cacheability} (cache-hit probability) as
\begin{equation}
p_{\text{cache}}
\;\triangleq\;
\frac{N_{\text{gen}}}{N_{\text{fwd}}}
\quad\in (0,1],
\label{eq:pcache}
\end{equation}
which can be interpreted as: during post-prefill decoding, a processed token instance becomes a final, cache-reusable token with probability $p_{\text{cache}}$. Equivalently, the \emph{average recomputation factor} is $1/p_{\text{cache}}$.

This metric reflects an efficiency dimension that is distinct from per-step parallelism. Fully bidirectional methods (e.g., LLaDA, Dream) may predict many tokens per forward, yet often achieve low $p_{\text{cache}}$ because few predictions are immediately cache-valid. Block-wise methods (e.g., SDAR, NBDiff) improve speed largely by increasing $p_{\text{cache}}$ via partial prefix commitment. Hence, in KV-cached decoding, improving $p_{\text{cache}}$ can match or exceed speedups from increasing per-step parallelism, making it a primary objective for inference-oriented decoding.

\paragraph{Implication.}
In most MDLM-style decoders, $p_{\text{cache}}$ collapses mainly due to two structural issues:
(i) \emph{bidirectional KV coupling}---token representations depend on future (unresolved) tokens, so even early predictions are not cache-valid and must be recomputed; and
(ii) \emph{out-of-order resolution}---later positions are often resolved before earlier ones, disrupting the left-to-right prefix structure that KV caching requires.
We address (i) by enforcing standard causal attention in \texttt{WeDLM}  (see \S\ref{sec:method}), and address (ii) with inference-time mechanisms that bias decoding toward left-to-right commitment (see \S\ref{sec:inference}).

\subsection{Rethinking the Necessity of Bidirectional Attention}
\label{subsec:rethinking_bidirectional}

Masked diffusion language models (MDLMs) recover masked tokens conditioned on the available (unmasked) context. Standard MDLMs~\citep{nie2025large,ye2025dream} typically adopt bidirectional attention so that each position can aggregate information from all others. While natural, this is \emph{not} a requirement of the mask-recovery objective itself.

Our key observation is that the \emph{information flow} needed for mask recovery can be realized under \emph{standard causal attention} by enforcing two algorithmic principles:

\begin{tcolorbox}[
    colback=teal!5!white,
    colframe=teal!60!black,
    arc=2mm,
    boxrule=1pt,
    left=4pt, right=4pt, top=4pt, bottom=4pt,
    title=\textbf{\textit{Design Principles for Causal Mask Recovery}}
]
Let $\mathcal{O}$ denote the set of observed (unmasked) positions and $\mathcal{M}$ the set of masked positions.
\begin{enumerate}[label=(\roman*), leftmargin=*, itemsep=2pt, topsep=2pt]
    \item \textbf{Observed-Context Visibility:} Each masked position should be able to attend to all observed tokens $x_{\mathcal{O}}$, enabling prediction with global observed context.
    \item \textbf{Directed Dependence Among Masks:} Mask-to-mask interactions need not be symmetric. We impose a directed ordering over $\mathcal{M}$ so that each masked position attends only to a subset of masked positions that precede it in the chosen order. This replaces within-step mutual visibility with a causal dependence structure implementable by causal attention.
\end{enumerate}
\end{tcolorbox}

Principle~(i) captures the essential requirement of MDLM-style denoising: masked predictions should be allowed to condition on all observed evidence.
Principle~(ii) is a modeling choice: we parameterize dependencies within the masked set using a causal factorization under an ordering $\pi$.
Concretely, we model the conditional joint as
$
q_\theta(x_\mathcal{M}\mid x_\mathcal{O};\pi)
=\prod_{j=1}^{|\mathcal{M}|} q_\theta\!\left(x_{\pi(j)} \mid x_\mathcal{O}, x_{\pi(<j)}\right).
$
This directed factorization allows earlier-resolved masked tokens to influence later ones without requiring bidirectional attention.

We do \emph{not} claim equivalence to bidirectional-attention MDLMs. Rather, the two principles above are sufficient to \emph{implement} set-conditioned masked prediction using standard causal attention. Whether the directed dependence in Principle~(ii) is sufficient in practice is empirical, which we evaluate in \S\ref{sec:experiments}.

\paragraph{Design requirement for \texttt{WeDLM}.}
Combining \S\ref{subsec:prefix_cacheability} and \S\ref{subsec:rethinking_bidirectional}, our goal is: keep \emph{standard causal attention} while ensuring that each masked position can access the \emph{full observed context}.
In the next section, we show how to satisfy this requirement via \textit{Topological Reordering}, and then address the training--inference gap using \textit{Dual-Stream Masking}.

\section{WeDLM Training: Causal Mask Recovery}
\label{sec:method}

This section presents the training framework of \texttt{WeDLM}, which reconciles parallel language decoding with \emph{standard causal attention}.
Building on the analysis in \S\ref{sec:analysis}, our training design targets two requirements:
(1) \emph{prefix-cache-compatible} computation under a strict causal mask, and
(2) \emph{full observed-context visibility} for mask recovery.
We first introduce \textit{Topological Reordering} (\S\ref{subsec:reordering}), which exposes the entire observed set to masked positions under an unmodified causal mask.
We then present \textit{Dual-Stream Masking} (\S\ref{subsec:dual_stream}) to mitigate the training--inference mismatch induced by prefix-conditioned decoding at inference time.

\subsection{Causal Mask Recovery via Topological Reordering}
\label{subsec:reordering}

To enforce the visibility constraint that masked positions attend to the full observed context under standard causal masking, we introduce \textit{Topological Reordering}. Instead of using bidirectional attention masks, we apply a permutation that places all observed tokens before masked tokens in the \emph{physical computation order}. We decouple physical order from \emph{logical text positions} (indexed by position embeddings), so that every masked token can attend to all observed tokens using unmodified causal attention.

\paragraph{Problem Setup.}
Consider a clean sequence $\mathbf{x}_0 = [x_1, x_2, \dots, x_L]$ with logical positions $\mathbf{p} = [1, 2, \dots, L]$. We sample a masking ratio $\gamma \in (0, 1]$ and uniformly select a subset of indices $\mathcal{M} \subset \{1, \dots, L\}$ with $|\mathcal{M}| = \gamma L$ to be masked. The remaining indices $\mathcal{O} = \{1, \dots, L\} \setminus \mathcal{M}$ are observed tokens, with $|\mathcal{O}| = N_o$ and $|\mathcal{M}| = N_m$.

\begin{figure}[t]
    \centering
        \includegraphics[width=0.98\textwidth]{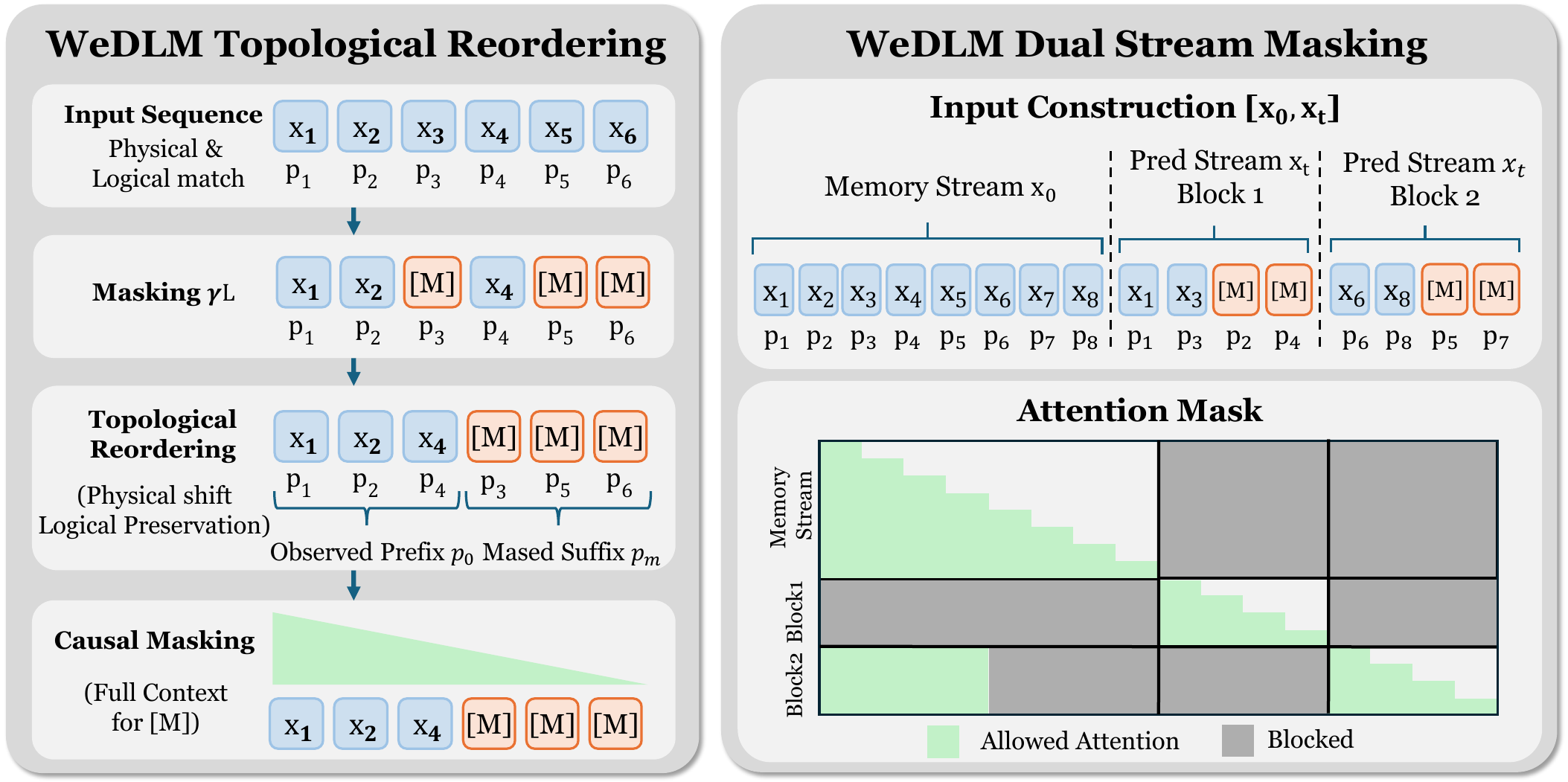}
        \caption{\textbf{Overview of the \texttt{WeDLM} training framework.} \textbf{Left:} \textit{Topological Reordering} physically shifts observed tokens to the prefix while preserving logical positions. This grants masked tokens access to the full observed context under standard causal masking. \textbf{Right:} \textit{Dual-Stream Masking} concatenates a clean Memory Stream with a masked Prediction Stream. The block-wise attention mask ensures that the Prediction Stream conditions on clean memory history rather than noisy preceding predictions, aligning training dynamics with inference.}
    \label{fig:wedlm-method}
\end{figure}

\paragraph{Topological Reordering.}
Standard MDLMs use bidirectional attention so that masked tokens can access observed tokens regardless of position. We provide the same \emph{observed-context visibility} under causal attention via a reordering operation. Specifically, we construct a reordered sequence $\tilde{\mathbf{x}}$ by placing all observed tokens before all masked tokens:
\begin{equation}
    \tilde{\mathbf{x}} = [\underbrace{x_{o_1}, x_{o_2}, \dots, x_{o_{N_o}}}_{\text{observed tokens}},\ \underbrace{\texttt{[M]}, \texttt{[M]}, \dots, \texttt{[M]}}_{N_m \text{ mask tokens}}],
    \label{eq:reordered_sequence}
\end{equation}
where $\{o_1, o_2, \dots, o_{N_o}\}$ are observed indices sorted in ascending order and \texttt{[M]} is a shared mask token. We preserve logical positions through a reordered position sequence:
\begin{equation}
    \tilde{\mathbf{p}} = [\underbrace{o_1, o_2, \dots, o_{N_o}}_{\mathbf{p}_o},\ \underbrace{m_1, m_2, \dots, m_{N_m}}_{\mathbf{p}_m}],
    \label{eq:reordered_positions}
\end{equation}
where $\{m_1, m_2, \dots, m_{N_m}\}$ are masked indices, also sorted in ascending order.

\paragraph{Context Awareness under Causal Masking.}
Under causal attention, a token at physical index $i$ can only attend to indices $\{1, \dots, i-1\}$. In $\tilde{\mathbf{x}}$, observed tokens occupy $\{1, \dots, N_o\}$ and masked tokens occupy $\{N_o+1, \dots, L\}$; thus every masked token can attend to \emph{all} observed tokens. Positional encodings (e.g., RoPE) are indexed by logical positions $\tilde{\mathbf{p}}$, so attention scores depend on logical relative offsets rather than physical indices.

\paragraph{Training Objective.}
With $(\tilde{\mathbf{x}}, \tilde{\mathbf{p}})$, we train the model to recover the ground-truth tokens at masked positions. For the $j$-th masked position (physical index $N_o + j$ with logical position $m_j$), the model predicts $x_0^{(m_j)}$ conditioned on the causal prefix. Following Eq.~\ref{eq:decoupled_lm}, we define:
\begin{equation}
    \mathcal{L}(\theta) = -\mathbb{E}_{\gamma, \mathbf{x}_0, \mathcal{M}} \left[ \frac{1}{\gamma} \sum_{j=1}^{N_m} \log P_\theta\left(x_0^{(m_j)} \mid \tilde{\mathbf{x}}_{<N_o+j},\ \tilde{\mathbf{p}}_{<N_o+j} \right) \right],
    \label{eq:wedlm_loss}
\end{equation}
where the factor $1/\gamma$ follows the weighting convention in Eq.~\ref{eq:mdlm_loss}. The key difference from bidirectional MDLMs is that we operate under strictly causal attention: each masked token conditions only on earlier \emph{physical} positions, yet still accesses the full observed context through topological reordering.

\subsection{Dual-Stream Masking for Training}
\label{subsec:dual_stream}

The objective in Eq.~\ref{eq:wedlm_loss} masks tokens uniformly over the sequence. During inference, however, decoding proceeds in a prefix-conditioned regime: the unresolved tokens predominantly reside in a (block-wise) suffix due to left-to-right progression, inducing a train--inference distribution gap. Related motivations appear in work bridging autoregressive and diffusion objectives~\citep{cheng2025sdar,tian2025next}. A naive fix---masking only short suffixes---would exclude most tokens from the loss. We therefore propose \textit{Dual-Stream Masking}, which simulates suffix-style decoding while preserving training efficiency.

\paragraph{Dual-Stream Construction.}
Given a clean sequence $\mathbf{x}_0 = [x_1, x_2, \dots, x_L]$ with positions $\mathbf{p} = [1, 2, \dots, L]$, we construct two copies: a \textit{memory stream} $\mathbf{x}_o$ and a \textit{prediction stream} $\mathbf{x}_t$, both initially identical to $\mathbf{x}_0$. These streams are concatenated to form the physical input:
\begin{equation}
    \mathbf{x}_{\text{input}} = [\underbrace{\mathbf{x}_o}_{\text{memory stream}},\ \underbrace{\mathbf{x}_t}_{\text{prediction stream}}].
    \label{eq:dual_stream_input}
\end{equation}
Critically, both streams share the same position sequence:
\begin{equation}
    \mathbf{p}_{\text{input}} = [\underbrace{1, 2, \dots, L}_{\mathbf{p}_o},\ \underbrace{1, 2, \dots, L}_{\mathbf{p}_t}].
    \label{eq:dual_stream_positions}
\end{equation}
This places the two streams in the same positional reference frame (e.g., under RoPE), enabling alignment between clean memory tokens and their masked counterparts in the prediction stream. The two streams are distinguished by their physical segment and the attention mask.

\paragraph{Block-wise Masking and Reordering.}
We partition the prediction stream $\mathbf{x}_t$ into $K = \lceil L / B \rceil$ non-overlapping blocks of size $B$ (except possibly the last). For each block $k \in \{1, \dots, K\}$, we sample a masking ratio $\gamma_k \in (0, 1]$ and apply masking, followed by \emph{intra-block} topological reordering within $\mathbf{x}_t$ as in \S\ref{subsec:reordering}: observed tokens in the block are moved to the front and masked tokens to the back, while logical positions are preserved. The memory stream $\mathbf{x}_o$ remains unmasked throughout and is not reordered.

\paragraph{Attention Pattern.}
The attention mask is designed to match inference-time conditioning: each block in $\mathbf{x}_t$ should rely on clean preceding context rather than noisy predictions. For a token in block $k$ of the prediction stream, its visible context is:
\begin{itemize}
    \item \textbf{Memory stream:} All memory tokens whose logical positions precede block $k$, providing clean context for earlier blocks.
    \item \textbf{Current block:} Tokens within block $k$ of $\mathbf{x}_t$ that precede the current token in the reordered physical sequence (standard causal masking).
\end{itemize}
Notably, tokens in block $k$ of $\mathbf{x}_t$ cannot attend to previous blocks within $\mathbf{x}_t$; they instead access the corresponding clean history from $\mathbf{x}_o$. This simulates the inference setting where earlier blocks are finalized and used as context.

\paragraph{Training Objective.}
Let $\mathcal{M}_k$ denote the set of masked logical positions within block $k$, and let $\tilde{\mathbf{x}}_t^{(k)}$ denote the reordered block in the prediction stream. We aggregate losses across blocks:
\begin{equation}
    \mathcal{L}(\theta) = -\mathbb{E}_{\{\gamma_k\}, \mathbf{x}_0} \left[ \sum_{k=1}^{K} \frac{1}{\gamma_k} \sum_{j \in \mathcal{M}_k} \log P_\theta\left(x_0^{(j)} \mid \mathbf{x}_o^{(<k)}, \tilde{\mathbf{x}}_t^{(k, <j)} \right) \right],
    \label{eq:dual_stream_loss}
\end{equation}
where $\mathbf{x}_o^{(<k)}$ denotes memory tokens whose logical positions precede block $k$, and $\tilde{\mathbf{x}}_t^{(k, <j)}$ denotes tokens in the reordered block $k$ that physically precede the masked position $j$. The factor $1/\gamma_k$ follows the per-block weighting convention.

\paragraph{Inference Compatibility.}
During inference, we discard the memory stream and decode with a standard causal attention mask over a single sequence. This requires no model-architecture changes and is compatible with optimized attention implementations such as FlashAttention~\cite{dao2022flashattention}.

\section{WeDLM Inference: Streaming Parallel Decoding}
\label{sec:inference}

In \S\ref{sec:analysis}, we introduced prefix cacheability $p_{\text{cache}}$ (Eq.~\ref{eq:pcache}) as a metric that captures \emph{how much of the post-prefill compute becomes reusable prefix KV states}.
In industrial KV-cached serving, latency is influenced not only by how many tokens are proposed per forward, but also by how effectively these parallel proposals can be \emph{committed into a contiguous left-to-right prefix}.
Here we instantiate this objective into a concrete procedure, \emph{Streaming Parallel Decoding}, which incrementally commits cache-ready tokens under standard causal attention while continuously refilling new masked positions to maintain steady GPU utilization.

\subsection{Inference Requirements for Streaming Decoding}
\label{subsec:inference_requirements}

Streaming Parallel Decoding is designed to maximize the rate of \emph{prefix commitment}.
After the initial prefill on a prompt prefix $\mathbf{x}$, the decoder repeatedly (i) predicts multiple masked positions in parallel, and (ii) converts sufficiently confident predictions into a \emph{committed prefix} whose KV states can be reused by subsequent steps.
To achieve a high $p_{\text{cache}}$ in practice, the inference procedure should satisfy the following requirements.

\begin{tcolorbox}[
    colback=orange!5!white,
    colframe=orange!60!black,
    arc=2mm,
    boxrule=1pt,
    left=4pt, right=4pt, top=4pt, bottom=4pt,
    title=\textbf{\textit{Inference Requirements for High Prefix Commitment}}
]
\begin{enumerate}[label=(\roman*), leftmargin=*, itemsep=2pt, topsep=2pt]
    \item \textbf{Immediate KV Validity (No Bidirectional Coupling).}
    Once a token is predicted, its KV representation should depend \emph{only} on already-committed context; otherwise it cannot be cached immediately and will require recomputation later, lowering $p_{\text{cache}}$.
    \item \textbf{Left-to-Right Commitment (Avoid Out-of-Order Resolution).}
    Since standard KV caching only supports a contiguous prefix, decoding should prioritize resolving earlier logical positions so that each step extends the committed prefix as much as possible.
    \item \textbf{No Stop-and-Wait (Avoid Pipeline Bubbles).}
    The algorithm should avoid stalling at block boundaries; it should maintain a steady workload per forward by introducing new masked positions as soon as tokens are committed.
\end{enumerate}
\end{tcolorbox}

\paragraph{How \texttt{WeDLM} meets these requirements.}
Requirement (i) is enabled by \texttt{WeDLM}'s \emph{standard causal attention}: under a strict causal mask, each token's KV state depends only on tokens at earlier \emph{physical} indices.
Combined with inference-time topological reordering that places already-resolved tokens before unresolved masks (while preserving logical positions via position ids), a resolved token becomes cache-valid \emph{immediately} after it is produced.
Requirements (ii)--(iii) are addressed by Streaming Parallel Decoding: we use a position-aware confidence rule that biases resolution toward the left, and a dynamic sliding window with on-the-fly refilling to remove block-boundary synchronization.

\subsection{Streaming Parallel Decoding}
\label{subsec:streaming_decoding}

We now describe \emph{Streaming Parallel Decoding}, an inference strategy that operates entirely under standard causal attention and KV caching.
The key idea is to maintain a fixed-size window of $W$ slots, containing a mixture of filled tokens (already predicted but not yet committed) and \texttt{[M]} tokens (pending prediction).
At each step, we: \textbf{(a) reorder} the window so that filled tokens appear before masks (positions preserved via per-token global position ids), \textbf{(b) run a causal forward} conditioned on the persistent cache, \textbf{(c) commit} the leftmost contiguous filled prefix (now cache-valid), \textbf{(d) predict} additional mask slots based on confidence and position, and \textbf{(e) refill} the window with new masks to keep parallelism constant.

\begin{figure}[t]
    \centering
        \includegraphics[width=0.98\textwidth]{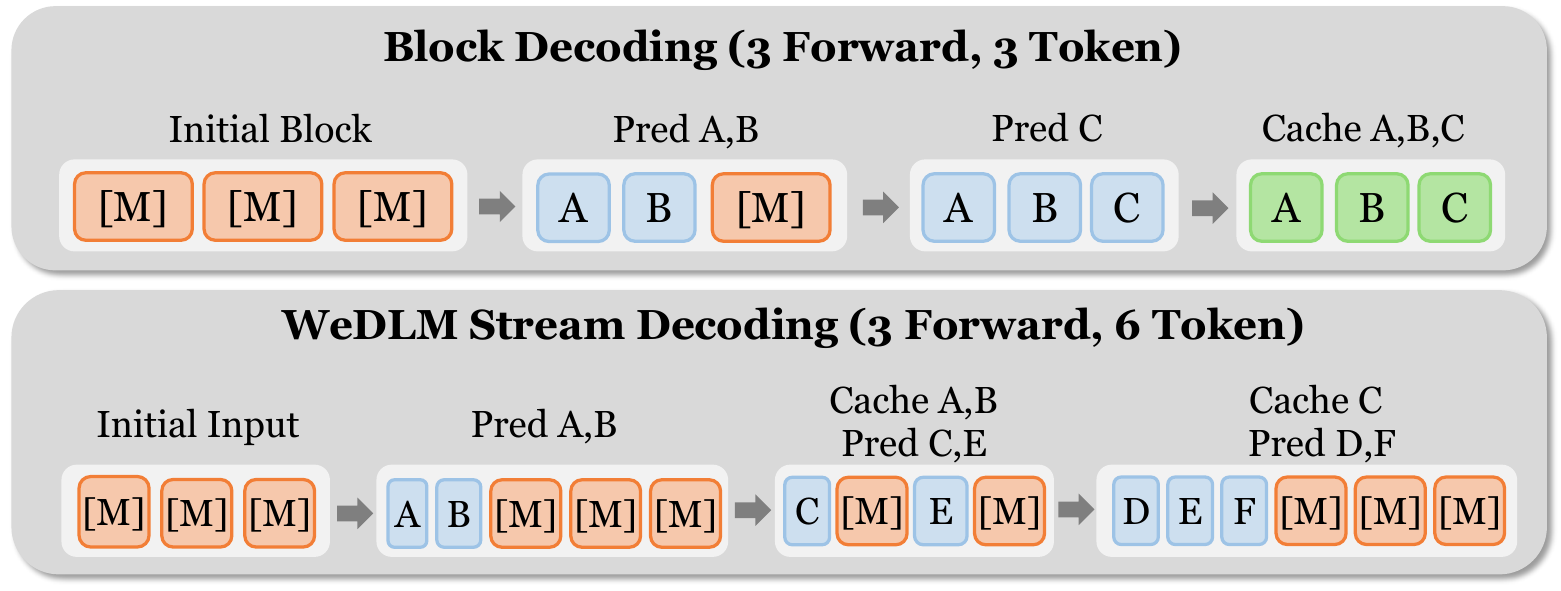}
        \caption{\textbf{Block Decoding vs.\ WeDLM Streaming Parallel Decoding.}
Block decoding suffers from stop-and-wait: bidirectional dependence within a block prevents committing any token until the entire block is finalized.
In contrast, \texttt{WeDLM} uses standard causal attention with a dynamic sliding window: resolved tokens (e.g., A, B) are immediately cache-ready and committed, while new mask tokens (e.g., C, E) are appended for parallel prediction.}
    \label{fig:wedlm-inference}
\end{figure}

\begin{algorithm}[t]
\caption{Streaming Parallel Decoding}
\label{alg:streaming_decoding}
\begin{algorithmic}[1]
\Require Prompt prefix $\mathbf{x}$, window size $W$, entropy threshold $\tau$, distance penalty $\lambda$
\Ensure Generated sequence $\mathbf{y}$
\State $\mathbf{y} \gets [\,]$; $(\mathbf{K}, \mathbf{V}) \gets \textsc{Prefill}(\mathbf{x})$
\State $\mathcal{W} \gets [\texttt{[M]}]^W$ \Comment{Each slot carries a fixed global position id}
\While{$\mathcal{W} \neq \emptyset$}
    \Statex \Comment{\textbf{Reorder \& Forward}: filled tokens placed before masks (logical positions preserved)}
    \State $\mathcal{W} \gets [\mathcal{W}_{\text{filled}}; \mathcal{W}_{\text{mask}}]$
    \State $(\boldsymbol{\ell}, \mathbf{K}_{\mathcal{W}}, \mathbf{V}_{\mathcal{W}}) \gets \textsc{Forward}(\mathcal{W}, \mathbf{K}, \mathbf{V})$

    \Statex \Comment{\textbf{Commit}: commit the leftmost contiguous filled prefix (cache-valid under causal mask)}
    \State $n \gets \min\{i : \mathcal{W}[i] = \texttt{[M]}\}$ or $|\mathcal{W}|$ if none
    \State Append $\mathcal{W}[0{:}n]$ to $\mathbf{y}$
    \State Extend $(\mathbf{K}, \mathbf{V})$ with $(\mathbf{K}_{\mathcal{W}}[0{:}n], \mathbf{V}_{\mathcal{W}}[0{:}n])$
    \State $\mathcal{W} \gets \mathcal{W}[n{:}]$

    \Statex \Comment{\textbf{Predict}: fill a subset of masks based on confidence and position bias}
    \State $\mathcal{F} \gets \textsc{SelectByEntropy}(\boldsymbol{\ell}_{\text{mask}}, \tau, \lambda)$
    \State $\mathcal{W}[i] \gets \textsc{Sample}(\boldsymbol{\ell}_i)$ for $i \in \mathcal{F}$

    \Statex \Comment{\textbf{Refill}: append new masks to maintain constant parallelism}
    \State $\mathcal{W} \gets [\mathcal{W}; [\texttt{[M]}]^n]$
\EndWhile
\State \Return $\mathbf{y}$
\end{algorithmic}
\end{algorithm}

\paragraph{Distance Penalty for Left-to-Right Commitment.}
To increase the chance that resolved tokens form a long contiguous prefix, we bias mask selection toward earlier positions.
Following~\citet{ye2025dream}, we use prediction entropy to determine which masks to fill.
Let $p_i(\cdot)$ be the predicted distribution at mask slot $i$ with entropy $H_i = -\sum_v p_i(v)\log p_i(v)$.
We define a distance-adjusted entropy:
\begin{equation}
    \tilde{H}_i = H_i + \lambda \cdot d_i,
    \label{eq:adjusted_entropy}
\end{equation}
where $d_i$ is the distance from slot $i$ to the leftmost remaining mask slot in the current window, and $\lambda>0$ controls the strength of left-to-right preference.
\textsc{SelectByEntropy} returns mask indices whose adjusted entropy falls below a threshold $\tau$.
This reduces out-of-order resolution patterns and accelerates contiguous prefix growth, improving $p_{\text{cache}}$.

\paragraph{Immediate Caching under Standard Causal Attention.}
Under \texttt{WeDLM}'s strict causal mask, the KV representation of a token depends only on the physical prefix.
After reordering, when filled tokens occupy the leftmost slots of $\mathcal{W}$, they attend only to the persistent cached prefix and earlier filled slots.
Therefore, the leftmost contiguous filled prefix is \emph{immediately cache-valid} and can be committed to $(\mathbf{K},\mathbf{V})$ without waiting for other mask slots to be resolved.
This directly addresses the bidirectional KV-coupling issue highlighted in \S\ref{sec:analysis}.

\paragraph{Dynamic Sliding Window to Eliminate Stop-and-Wait.}
Block-wise decoders must wait until an entire block becomes final before committing, creating pipeline bubbles.
In contrast, streaming maintains a fixed window size $W$.
At each step, committed tokens are removed and an equal number of new \texttt{[M]} slots are appended.
This on-the-fly refill keeps the amount of work per forward approximately constant, maintaining computational saturation and avoiding block-boundary synchronization.

\paragraph{Compatibility with Efficient Inference Systems.}
Streaming Parallel Decoding derives its efficiency from operating entirely under standard causal attention.
Each decoding step reduces to a causal forward over the current window conditioned on the cached prefix---effectively a small prefill---which is natively supported by FlashAttention~\citep{dao2022flashattention}, PagedAttention~\citep{vllm}, and CUDA Graphs without kernel modification.
This design enables direct deployment on industrial AR inference infrastructure.

\section{Experiments}
\label{sec:experiments}

This section evaluates \texttt{WeDLM} across multiple dimensions. We first describe the experimental setup (\S\ref{subsec:exp_setup}), then present results on generation quality (\S\ref{subsec:quality_results}) and inference efficiency (\S\ref{subsec:efficiency_results}), followed by ablation studies (\S\ref{subsec:ablation}).

\subsection{Training Details}
\label{subsec:exp_setup}

We initialize \texttt{WeDLM} from pre-trained autoregressive models in the Qwen family. Specifically, we use Qwen2.5-7B~\cite{qwen2.5} and Qwen3-8B~\cite{qwen3} as our base models. These models provide strong foundations with well-established performance across diverse tasks.
We perform continued pretraining on 100B tokens to adapt the base models to the \texttt{WeDLM} framework. The learning rate starts at $3 \times 10^{-6}$ and gradually decays to $3 \times 10^{-7}$ following a cosine schedule. For the Dual-Stream Masking strategy described in \S\ref{subsec:dual_stream}, we set the block size $B = 32$. To efficiently handle the irregular attention patterns introduced by topological reordering, we employ Magi Attention~\cite{magi}, which accelerates computation over non-rectangular attention masks without requiring custom CUDA kernels.
To preserve the original autoregressive capabilities, we incorporate an auxiliary AR loss during training. This loss is computed on the same sequences using standard next-token prediction, ensuring that the model retains its causal language modeling ability while learning the masked diffusion objective.
After pretraining, we perform supervised fine-tuning (SFT) to improve instruction-following capabilities. We use 10K internal instruction-response pairs for this stage. The learning rate is set to $3 \times 10^{-6}$ with a cosine decay schedule.
The resulting models are denoted as \texttt{WeDLM-7B} (based on Qwen2.5-7B) and \texttt{WeDLM-8B} (based on Qwen3-8B).

\begin{table}[t]
\centering
\caption{Main results on generation quality across diverse benchmarks for \textbf{Base models}. We compare our \texttt{WeDLM} against autoregressive (AR) baselines and recent diffusion language models (DLLMs). The columns for our model are highlighted in blue. Best results in each row are \textbf{bolded}.}
\label{tab:base_results}
\definecolor{highlightblue}{RGB}{235, 245, 255}
\definecolor{headergray}{RGB}{242, 242, 242}

\resizebox{\textwidth}{!}{%
\begin{tabular}{@{}l|cc|cc|>{\columncolor{highlightblue}}c>{\columncolor{highlightblue}[\tabcolsep][0pt]}c@{}}
\toprule
\multirow{3}{*}{\textbf{Benchmark}} & \multicolumn{2}{c|}{\textbf{AR Baseline}} & \multicolumn{2}{c|}{\textbf{DLLM Baseline}} & \multicolumn{2}{c}{\cellcolor{highlightblue}\textbf{WeDLM (Ours)}} \\
  & \multicolumn{2}{c|}{\textit{Base Model}} & \multicolumn{2}{c|}{\textit{Base Model}} & \multicolumn{2}{c}{\cellcolor{highlightblue}\textit{Base Model}} \\
\cmidrule(lr){2-3} \cmidrule(lr){4-5} \cmidrule(lr){6-7}
  & Qwen2.5-7B & Qwen3-8B & LLaDA-8B & Dream-7B & \textbf{\texttt{WeDLM-7B}} & \textbf{\texttt{WeDLM-8B}} \\ 
\midrule

% General Tasks Section
\rowcolor{headergray}
\multicolumn{7}{c}{\textbf{\textit{General Reasoning}}} \\ 
\midrule
ARC-C \scriptsize{(0-shot)} & 89.93 & 92.66 & 81.14 & 88.40 & 90.70 & \textbf{92.92} \\
ARC-E \scriptsize{(0-shot)} & 96.55 & 97.13 & 92.00 & 96.21 & 96.13 & \textbf{97.14} \\
HellaSwag \scriptsize{(10-shot)} & 80.20 & 85.27 & \textbf{85.34} & 78.05 & 85.11 & 84.55 \\
MMLU \scriptsize{(5-shot)} & 71.62 & 74.03 & 64.61 & 70.64 & 71.93 & \textbf{75.46} \\
\midrule
\addlinespace[0.4em]

% Math & Science Section
\rowcolor{headergray}
\multicolumn{7}{c}{\textbf{\textit{Math \& Science}}} \\ 
\midrule
GSM8K \scriptsize{(3-shot)} & 79.23 & 85.97 & 71.80 & 75.97 & 84.76 & \textbf{90.20} \\
MATH \scriptsize{(4-shot)} & 43.40 & 50.80 & 28.00 & 38.00 & 48.20 & \textbf{53.60} \\
GPQA-Diamond \scriptsize{(5-shot)} & 33.70 & 37.00 & 29.80 & 25.76 & 36.87 & \textbf{42.42} \\
\midrule
\addlinespace[0.4em]

% Code Section
\rowcolor{headergray}
\multicolumn{7}{c}{\textbf{\textit{Code Generation}}} \\ 
\midrule
MBPP \scriptsize{(3-shot)} & 65.30 & \textbf{70.94} & 41.99 & 56.47 & 61.81 & 67.00 \\
HumanEval \scriptsize{(4-shot)} & 59.14 & 68.90 & 31.71 & 20.12 & 68.90 & \textbf{75.00} \\
HumanEval-plus \scriptsize{(4-shot)} & 53.05 & 63.40 & 28.05 & 19.51 & 64.00 & \textbf{68.90} \\
\midrule
\addlinespace[0.4em]

% Average Section
\textbf{Average} & 67.21 & 72.61 & 55.44 & 56.91 & 70.84 & \textbf{74.72} \\
\bottomrule
\end{tabular}%
}
\end{table}

\subsection{Evaluation Setup}
\label{subsec:eval_setup}

We evaluate \texttt{WeDLM} on a diverse set of benchmarks spanning reasoning, knowledge, and code generation. For knowledge and commonsense reasoning, we use ARC-Challenge~\cite{clark2018think} (0-shot), GPQA~\cite{rein2024gpqa} (5-shot), HellaSwag~\cite{zellers2019hellaswag} (10-shot), and MMLU~\cite{hendrycks2021mmlu} (5-shot). For mathematical reasoning, we evaluate on GSM8K~\cite{cobbe2021training} (3-shot) and MATH~\cite{hendrycks2020measuring} (4-shot). For code generation, we use MBPP~\cite{austin2021program} (3-shot) and HumanEval~\cite{chen2021evaluating} (4-shot). For generative tasks (GSM8K, GPQA, MBPP, HumanEval, and MATH), we set the maximum generation length to 512 tokens and use a sampling temperature of 0.1. For inference, to ensure fair comparison, the results in Tables~\ref{tab:base_results} and~\ref{tab:instruct_results} are obtained with a unified step-wise decoding scheme: at each step, all methods (including our model and diffusion baselines) generate one token by selecting the position with the lowest entropy; unless otherwise specified, we use a window size $W=6$ and a distance-based penalty coefficient $\lambda=0.10$ (see Eq.~\ref{eq:adjusted_entropy}) to balance generation quality and speed. We compare \texttt{WeDLM} against both autoregressive baselines and recent diffusion language models. The autoregressive baselines include Qwen2.5-7B and Qwen3-8B, which serve as the base models for our method. For diffusion models, we compare against LLaDA-8B~\cite{nie2025large}, Dream-7B~\cite{ye2025dream}, and SDAR-8B~\cite{cheng2025sdar}. To ensure fair comparison, each model uses its recommended inference framework: LLaDA and Dream use dInfer, SDAR uses JetEngine, and the Qwen models use vLLM~\cite{vllm}. Our \texttt{WeDLM} models are also served via vLLM, demonstrating seamless compatibility with industrial inference systems.

\begin{table}[t]
\centering
\caption{Main results on generation quality across diverse benchmarks for \textbf{Instruct models}. We compare our \texttt{WeDLM} against autoregressive (AR) baselines and recent diffusion language models (DLLMs). The columns for our model are highlighted in blue. Best results in each row are \textbf{bolded}.}
\label{tab:instruct_results}
\definecolor{highlightblue}{RGB}{235, 245, 255}
\definecolor{headergray}{RGB}{242, 242, 242}

\resizebox{\textwidth}{!}{%
\begin{tabular}{@{}l|cc|ccc|>{\columncolor{highlightblue}}c>{\columncolor{highlightblue}[\tabcolsep][0pt]}c@{}}
\toprule
\multirow{3}{*}{\textbf{Benchmark}} & \multicolumn{2}{c|}{\textbf{AR Baseline}} & \multicolumn{3}{c|}{\textbf{DLLM Baseline}} & \multicolumn{2}{c}{\cellcolor{highlightblue}\textbf{WeDLM (Ours)}} \\
  & \multicolumn{2}{c|}{\textit{Instruct Model}} & \multicolumn{3}{c|}{\textit{Instruct Model}} & \multicolumn{2}{c}{\cellcolor{highlightblue}\textit{Instruct Model}} \\
\cmidrule(lr){2-3} \cmidrule(lr){4-6} \cmidrule(lr){7-8}
  & Qwen2.5-7B & Qwen3-8B & LLaDA-8B & Dream-7B & SDAR-8B & \textbf{\texttt{WeDLM-7B}} & \textbf{\texttt{WeDLM-8B}} \\ 
\midrule

% General Tasks Section
\rowcolor{headergray}
\multicolumn{8}{c}{\textbf{\textit{General Reasoning}}} \\ 
\midrule
ARC-C \scriptsize{(0-shot)} & 86.09 & 91.47 & 85.92 & 87.20 & 91.13 & 89.59 & \textbf{92.92} \\
ARC-E \scriptsize{(0-shot)} & 93.27 & 96.17 & 94.32 & 93.27 & 97.01 & 96.09 & \textbf{97.43} \\
HellaSwag \scriptsize{(10-shot)} & 87.59 & 86.13 & 78.55 & 62.00 & \textbf{92.12} & 84.75 & 82.94 \\
MMLU \scriptsize{(5-shot)} & 71.98 & 71.52 & 63.70 & 64.19 & 73.61 & 70.52 & \textbf{75.14} \\
\midrule
\addlinespace[0.4em]

% Math & Science Section
\rowcolor{headergray}
\multicolumn{8}{c}{\textbf{\textit{Math \& Science}}} \\ 
\midrule
GSM8K \scriptsize{(3-shot)} & 89.91 & 89.91 & 80.59 & 79.00 & 91.66 & 87.57 & \textbf{92.27} \\
MATH \scriptsize{(4-shot)} & 45.00 & \textbf{69.60} & 34.20 & 41.00 & 43.40 & 55.40 & 64.80 \\
GPQA-Diamond \scriptsize{(5-shot)} & 27.10 & 41.41 & 25.25 & 35.86 & 38.38 & 33.84 & \textbf{44.95} \\
\midrule
\addlinespace[0.4em]

% Code Section
\rowcolor{headergray}
\multicolumn{8}{c}{\textbf{\textit{Code Generation}}} \\ 
\midrule
MBPP \scriptsize{(3-shot)} & 63.66 & 68.37 & 36.24 & 58.52 & 67.97 & 63.66 & \textbf{70.53} \\
HumanEval \scriptsize{(4-shot)} & 76.22 & 71.95 & 36.59 & 57.32 & 76.83 & 75.00 & \textbf{80.49} \\
HumanEval-plus \scriptsize{(4-shot)} & 70.12 & 64.63 & 32.32 & 51.22 & 70.12 & 71.34 & \textbf{73.78} \\
\midrule
\addlinespace[0.4em]

% Average Section
\textbf{Average} & 71.09 & 75.12 & 56.77 & 62.96 & 74.22 & 72.78 & \textbf{77.53} \\
\bottomrule
\end{tabular}%
}
\end{table}

\subsection{Performance Evaluation}
\label{subsec:quality_results}

Tables~\ref{tab:base_results} and \ref{tab:instruct_results} report generation quality for \texttt{WeDLM} under \emph{base} and \emph{instruct} settings, respectively. Across both settings, the main trend is consistent: \texttt{WeDLM} not only preserves but often improves upon the capabilities of its underlying autoregressive (AR) checkpoints, while maintaining a large margin over prior diffusion language models.

On \textbf{base models} (Table~\ref{tab:base_results}), \texttt{WeDLM-7B} achieves an average score of 70.84, improving over Qwen2.5-7B (67.21) by 3.6 points, and \texttt{WeDLM-8B} reaches 74.72, exceeding Qwen3-8B (72.61) by 2.1 points. The gains concentrate on reasoning-heavy tasks: on GSM8K, \texttt{WeDLM-7B} improves by 5.5 points (84.76 vs.\ 79.23) and \texttt{WeDLM-8B} by 4.2 points (90.20 vs.\ 85.97). We observe similar improvements on MATH (+4.8 and +2.8 points) and GPQA-Diamond (+3.2 and +5.4 points). For code, \texttt{WeDLM} shows notable gains on HumanEval (+9.8 for 7B and +6.1 for 8B), while MBPP is the only benchmark with a consistent drop (about 3--4 points), suggesting sensitivity to domain or prompt-format differences.

On \textbf{instruct models} (Table~\ref{tab:instruct_results}), \texttt{WeDLM} remains competitive with, and in many cases surpasses, its AR instruct baselines. \texttt{WeDLM-7B} improves over Qwen2.5-7B on ARC-C (+4.0), ARC-E (+2.9), MATH (+9.2), and GPQA-Diamond (+8.8), but underperforms on GSM8K (-3.3) and HumanEval (-3.1). \texttt{WeDLM-8B} shows the strongest overall results, reaching an average of 77.53, which is +2.4 over Qwen3-8B (75.12). It delivers consistent gains on reasoning and code, including MMLU (+3.6), GPQA-Diamond (+3.5), HumanEval (+8.5), and HumanEval-plus (+9.2), while remaining close on GSM8K (+2.4) and MBPP (+2.2). These results indicate that the diffusion-style training objective and parallel decoding do not conflict with instruction tuning, and can even amplify it when starting from a strong instruct checkpoint.

Compared to \textbf{diffusion baselines}, \texttt{WeDLM} maintains a clear advantage in both settings. On base models, LLaDA-8B and Dream-7B average 55.44 and 56.91, which is 15--19 points below \texttt{WeDLM}. On instruct models, diffusion baselines improve but still lag behind: the best diffusion instruct baseline (SDAR-8B) averages 74.22, while \texttt{WeDLM-8B} reaches 77.53. The gap is most visible on code generation and higher-difficulty reasoning (e.g., HumanEval and GPQA-Diamond), where \texttt{WeDLM-8B} sets the best overall scores among the compared models.

\begin{figure}[t]
    \centering
    \includegraphics[width=\linewidth]{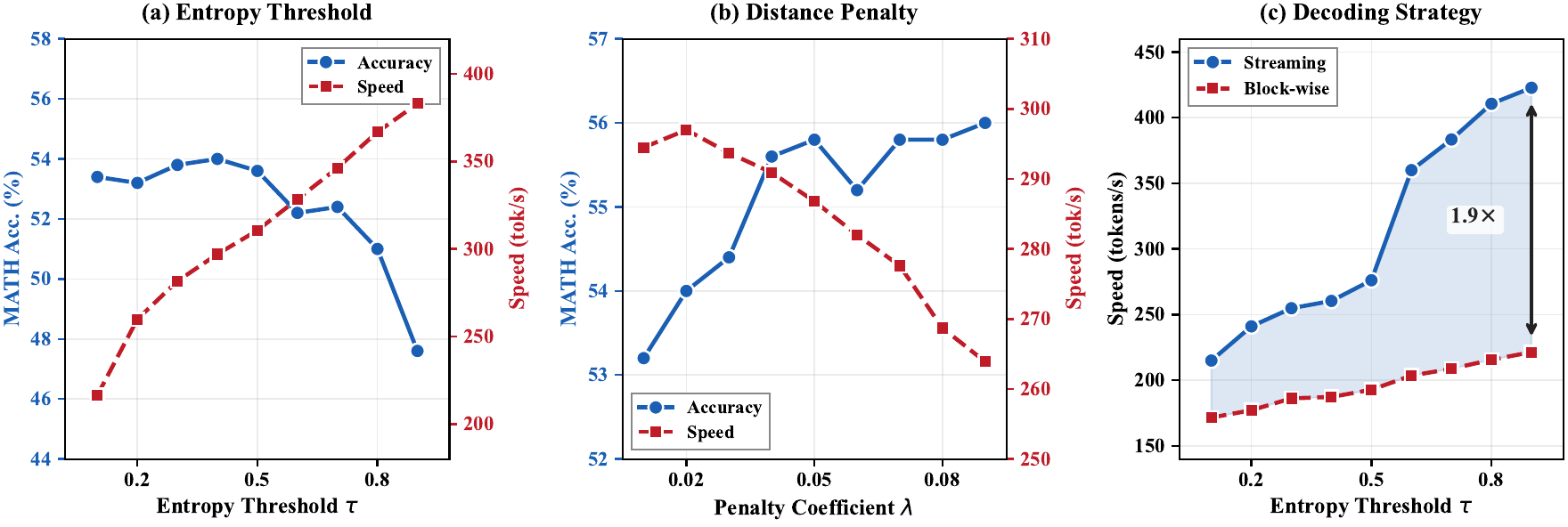}
    \caption{Ablation studies on inference hyperparameters. (a) Effect of entropy threshold $\tau$ on MATH accuracy and generation speed, revealing a quality-speed trade-off with optimal range $\tau \in [0.3, 0.5]$. (b) Effect of distance penalty coefficient $\lambda$, showing that prioritizing left-positioned tokens improves accuracy with minimal speed cost. (c) Comparison of Streaming Parallel Decoding versus block-wise decoding across entropy thresholds; streaming achieves up to $1.9\times$ speedup by enabling immediate prefix commitment.}
    \label{fig:speed_evaluation}
\end{figure}

\subsection{Speed Evaluation}
\label{subsec:efficiency_results}

\paragraph{Hyperparameter Sensitivity.}
Figure~\ref{fig:speed_evaluation} examines the key inference hyperparameters using \texttt{WeDLM-7B-Instruct}. The entropy threshold $\tau$ (Figure~\ref{fig:speed_evaluation}a) controls unmasking confidence: lower values yield higher accuracy but slower generation. Performance remains stable ($\sim$53--54\%) for $\tau \leq 0.5$, then degrades sharply at higher thresholds as low-confidence predictions propagate errors. We recommend $\tau \in [0.3, 0.6]$ for balanced operation. The distance penalty $\lambda$ (Figure~\ref{fig:speed_evaluation}b) biases selection toward left-positioned tokens. From the perspective of prefix cacheability (\S\ref{subsec:prefix_cacheability}), prioritizing earlier positions directly increases $p_{\text{cache}}$: tokens resolved earlier in the sequence are more likely to form a contiguous committed prefix, whose KV states become immediately reusable. Increasing $\lambda$ from 0.01 to 0.05 improves accuracy by 2.6 points with only 3\% speed reduction, confirming that left-to-right resolution not only accelerates caching but also reduces error accumulation from out-of-order predictions.

\paragraph{Streaming vs.\ Block-wise Decoding: A $p_{\text{cache}}$ Perspective.}
Figure~\ref{fig:speed_evaluation}(c) demonstrates that Streaming Parallel Decoding consistently outperforms block-wise decoding across all entropy thresholds. At $\tau = 0.9$, streaming achieves $1.9\times$ speedup (423 vs.\ 221 tokens/s). This gap can be understood through the lens of prefix cacheability (Eq.~\ref{eq:pcache}): block-wise methods must wait until an entire block is finalized before any token becomes cache-valid, yielding lower $p_{\text{cache}}$ due to synchronization overhead. In contrast, streaming commits tokens as soon as they form a contiguous left-to-right prefix, maximizing $p_{\text{cache}}$ by converting each resolved token into a cache-reusable state without delay.

\begin{figure}[t]
    \centering
    \includegraphics[width=\linewidth]{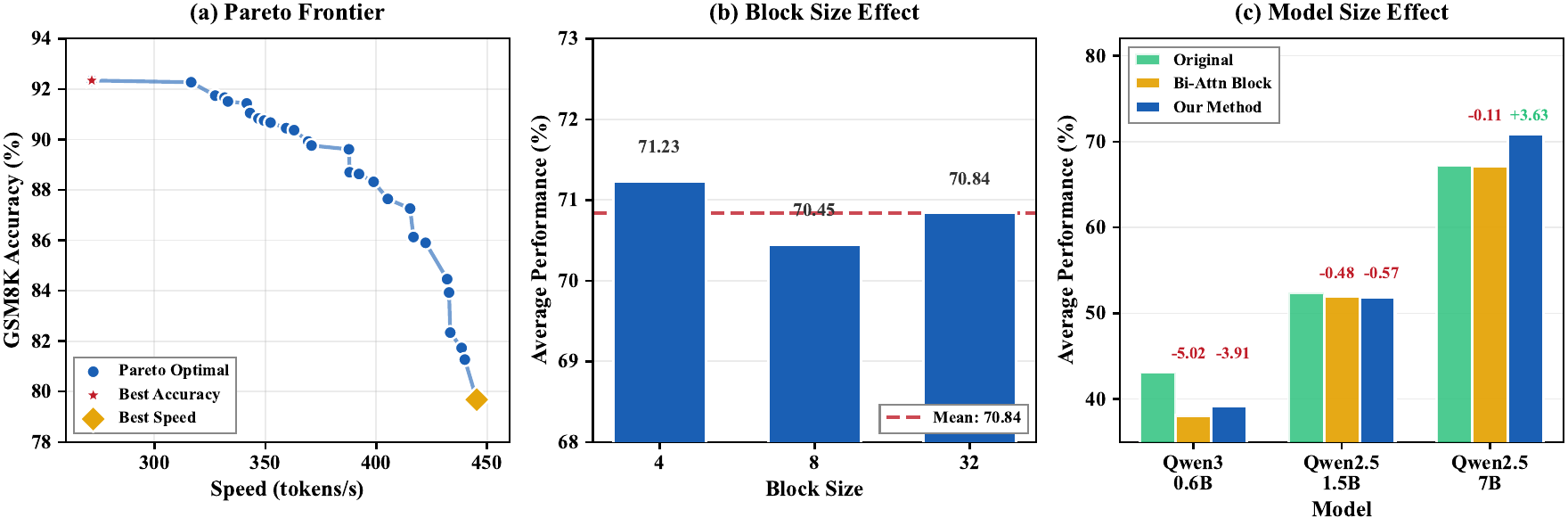}
    \caption{\textbf{Ablation studies.} (a) Pareto frontier on GSM8K showing quality-speed trade-offs across hyperparameter configurations; conservative settings achieve 92.3\% accuracy at $1.97\times$ speedup while aggressive settings reach $3.2\times$ acceleration. (b) Block size effect during continued pretraining shows stable performance across $B \in \{4, 8, 32\}$. (c) Attention design and model scale: we compare bidirectional attention within blocks (Bi-Attn Block) against our causal design (Our Method) across model sizes; larger models benefit more from causal adaptation, while bidirectional intra-block attention consistently underperforms.}
    \label{fig:ablation}
\end{figure}

\paragraph{Quality-Speed Pareto Frontier.}
Figure~\ref{fig:ablation}(a) presents the Pareto optimal configurations on GSM8K using \texttt{WeDLM-8B-Instruct}, spanning accuracy from 79.7\% to 92.3\% at speeds of 272--445 tokens/s. The frontier reveals a smooth trade-off: conservative settings ($\tau=0.2$, $\lambda=0.01$) preserve near-baseline accuracy (92.3\%) at $1.97\times$ speedup, while aggressive settings ($\tau=0.9$, $\lambda=0.01$) achieve $3.2\times$ acceleration with accuracy above 79\%. This flexibility allows practitioners to select operating points based on task-specific requirements.

\subsection{Ablation Studies}
\label{subsec:ablation}

\paragraph{Block Size.}
Figure~\ref{fig:ablation}(b) examines the effect of block size $B$ during continued pretraining. Performance remains virtually identical across $B \in \{4, 8, 32\}$, with average scores within a 0.8-point range (70.45--71.23), demonstrating that \texttt{WeDLM} is insensitive to block size. This flexibility favors larger block sizes in practice: libraries such as Magi Attention incur higher overhead for smaller blocks, and models trained with larger $B$ naturally support any smaller window size at inference time without retraining, providing greater deployment flexibility.

\paragraph{Causal vs.\ Bidirectional Intra-Block Attention.}
Figure~\ref{fig:ablation}(c) compares our fully causal design against a variant that uses bidirectional attention \emph{within} each prediction block (while remaining causal across blocks). This relaxes Principle~(ii) from \S\ref{subsec:rethinking_bidirectional} by allowing mutual visibility among masked tokens inside a block. Overall, causal intra-block attention achieves higher average performance than the bidirectional variant, indicating that the directed factorization is already sufficient for AR-initialized models. Moreover, bidirectional intra-block attention fundamentally limits $p_{\text{cache}}$---tokens cannot be committed until the entire block resolves---whereas our causal design enables immediate per-token caching.

\paragraph{Base Model Initialization.}
Figure~\ref{fig:ablation}(c) also investigates how model scale affects adaptation to the \texttt{WeDLM} framework. Smaller models (0.6B, 1.5B) experience slight performance degradation ($-3.9$ and $-0.6$ points for our method), while larger models (7B) show consistent improvements ($+3.6$ points). Notably, the improvement magnitude correlates monotonically with the base model's original capability: stronger AR checkpoints adapt more readily to the diffusion objective. This trend hints at a potential scaling law for AR-to-diffusion adaptation, where the benefit of diffusion training increases predictably with model capacity. We leave systematic verification of this hypothesis to future work, but these results already suggest that 7B+ scale models are the recommended starting point for \texttt{WeDLM} deployment.

\subsection{Case Study}
\label{subsec:case_study}

To better understand the performance characteristics of \texttt{WeDLM}, we analyze its generation behavior across different task modalities. The decoding speed is strongly correlated with the entropy of the output distribution, as shown in the representative cases in Appendix~\ref{sec:appendix_cases}:
\vspace{-0.5em} 
\begin{itemize}
    \setlength\itemsep{0pt} 
    \setlength\parskip{0pt}
    \setlength\parsep{0pt}
    
    \item \textbf{Low Entropy (Sequential Patterns):} As shown in Figure~\ref{fig:case_count}, the model achieves a peak throughput of \textbf{1673.3 tokens/s} on a simple counting task. The deterministic nature of the sequence yields extremely low entropy, allowing the model to speculate and accept many tokens per step.
    
    \item \textbf{Medium Entropy (Structured Reasoning):} Figure~\ref{fig:case_math} demonstrates a mathematical derivation task. Despite requiring logic, the syntactic structure of the solution is relatively predictable, maintaining a high speed of \textbf{745.2 tokens/s}.
    
    \item \textbf{High Entropy (Open-ended Generation):} In Figure~\ref{fig:case_qa}, where the model explains Quantum Physics, the speed drops to \textbf{197.8 tokens/s}. The high semantic diversity and lexical uncertainty in open-ended text reduce the confidence of speculative tokens, limiting the effective parallel block size.
\end{itemize}
\vspace{-0.5em} 

These results highlight a significant performance disparity: while low-entropy tasks achieve over $8\times$ speedup, high-entropy generation sees diminishing returns. Although this variance partially reflects the intrinsic uncertainty of natural language, it exposes a limitation of the current framework in handling high-perplexity scenarios. Closing this gap—potentially through more robust acceptance mechanisms or dynamic entropy calibration—remains a critical direction for future work to ensure consistent acceleration across all domains.

\section{Related Work}

\paragraph{Discrete Diffusion Language Models.}
Discrete diffusion models learn to iteratively denoise corrupted sequences, enabling parallel generation and bidirectional context modeling. RADD~\citep{ou2024your} simplified the framework by deriving a time-independent formulation of the concrete score, eliminating the need for time embeddings and enabling efficient caching. \citet{nie2024scaling} established scaling laws showing that while masked diffusion models require approximately 16$\times$ more compute to match autoregressive (AR) perplexity, they exhibit similar scaling trends. LLaDA~\citep{nie2025large} was the first to scale masked diffusion to 8B parameters, demonstrating competitive performance with AR models. LLaDA-MoE~\citep{zhu2025llada} further showed that sparse mixture-of-experts integrates effectively with masked diffusion, matching dense model performance with 1/6 active parameters. Recent work has also demonstrated that diffusion language models can be enhanced through reinforcement learning to improve reasoning capabilities~\citep{zhao2025d1,wang2025d2,pan2025d}.

\paragraph{Adapting Autoregressive Models to Diffusion.}
Given the substantial investment in pretrained AR models, recent works explore efficient adaptation strategies. DiffuLLaMA~\citep{gong2024scaling} introduced the shift operation to preserve AR's next-token prediction structure and attention mask annealing for gradual transition to bidirectional attention. Dream 7B~\citep{ye2025dream} proposed context-adaptive noise rescheduling to weight loss by contextual information density, achieving strong performance with only 0.6T tokens from Qwen2.5. Dream-Coder~\citep{xie2025dream} extended this approach to code generation, revealing emergent non-linear generation patterns such as sketch-first reasoning.

\paragraph{Block Diffusion and Inference Acceleration.}
Block diffusion methods apply diffusion within fixed-size blocks while maintaining AR dependencies across blocks. BD3-LM~\citep{arriola2025block} introduced vectorized training and clipped noise schedules to address gradient variance. NBDiff~\citep{tian2025next} viewed AR as block diffusion with block size 1 and proposed gradual block growth with context-causal attention for smooth adaptation. SDAR~\citep{cheng2025sdar} demonstrated lightweight AR-to-block-diffusion adaptation with dynamic confidence-based truncation, preserving model capabilities while enabling parallel decoding. Efficient-DLM~\citep{fu2025efficient} proposed block-wise attention that remains causal across blocks while enabling bidirectional modeling within each block, combined with position-dependent masking for effective AR-to-dLM conversion. SDLM~\citep{liu2025sequential} proposed adaptive-length speculative decoding using longest-prefix decoding within diffusion blocks. SBD~\citep{gat2025set} unified next-token and masked-token prediction within a single architecture, leveraging entropy-bounded samplers for flexible parallel decoding. LLaDA2.0~\citep{bie2025llada2} further demonstrated the scalability of block diffusion models to 100B parameters.

\paragraph{Permutation and Reordering.}
XLNet~\citep{yang2019xlnet} studies permutation language modeling, i.e., training an autoregressive objective under random factorization orders with permutation-dependent masking (via two-stream attention) to avoid information leakage.
\texttt{WeDLM} is different in both goal and mechanism: we focus on \emph{inference-time} acceleration with diffusion-style parallel decoding while \emph{keeping standard causal attention}.
Our Topological Reordering simply moves currently observed tokens to the physical prefix so masked tokens can attend to them under an unmodified lower-triangular mask, preserving logical positions (e.g., via RoPE position ids) and remaining KV-cache friendly.

\section{Conclusion}
\label{sec:conclusion}

We introduced \texttt{WeDLM}, a diffusion-style decoding framework that is explicitly optimized for \emph{prefix-cacheable} generation under \emph{standard causal attention}. Our analysis highlights that, in KV-cached decoding, the dominant efficiency driver is not merely ``tokens predicted per forward'', but the \emph{rate at which predictions become a contiguous left-to-right prefix} and therefore reusable, which we formalize via prefix cacheability $p_{\text{cache}}$ (Eq.~\ref{eq:pcache}). This viewpoint also clarifies why out-of-order resolution and bidirectional KV coupling are fundamentally misaligned with fast decoding: both reduce the fraction of computation that can be amortized by caching.

\texttt{WeDLM} addresses this mismatch by enforcing a causal dependency structure throughout training and inference. Topological Reordering exposes the full observed context to masked positions while preserving the strict causal mask, making each newly committed token immediately cache-valid. Building on this property, Streaming Parallel Decoding biases acceptance toward earlier logical positions and continuously refills a fixed window, converting parallel proposals into prefix growth with minimal recomputation. Empirically, \texttt{WeDLM} retains (and often improves) the capabilities of strong AR backbones while delivering substantial inference acceleration under matched, cache-enabled decoding.

More broadly, our results suggest that \emph{prefix-cacheability should be treated as a first-class design objective} for parallel text generation. Since the optimal reuse pattern is inherently close to prefix order, future diffusion language models should be constructed as more effective \emph{multi-token prediction (MTP)} mechanisms: generating many tokens per iteration is only beneficial insofar as those tokens can be quickly promoted into a cache-valid prefix under a causal computation graph. In this sense, causal diffusion provides a principled route to reconcile diffusion-style parallelism with the algorithmic structure required for efficient cached decoding.

% \newpage
% \bibliographystyle{plain}
\bibliography{main}

\newpage
\appendix

\section{Additional Qualitative Results}
\label{sec:appendix_cases}

\begin{figure*}[h]
    \centering
    \includegraphics[width=0.9\linewidth]{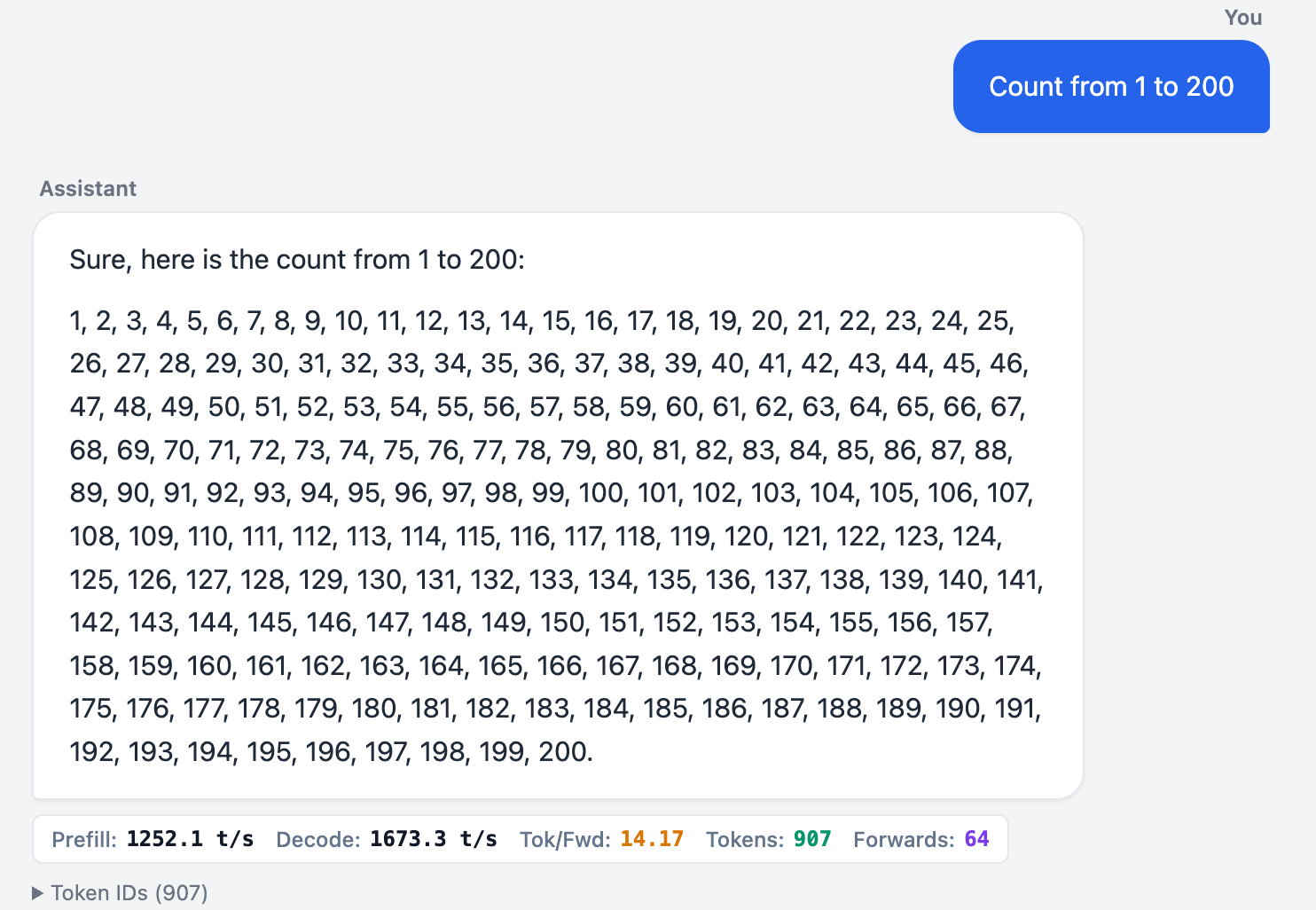}
    \caption{\textbf{Low Entropy Case:} A simple counting task from 1 to 200. Due to the highly predictable deterministic pattern, \texttt{WeDLM} achieves a decoding speed of \textbf{1673.3 tokens/s}.}
    \label{fig:case_count}
\end{figure*}

\begin{figure*}[h]
    \centering
    \includegraphics[width=0.9\linewidth]{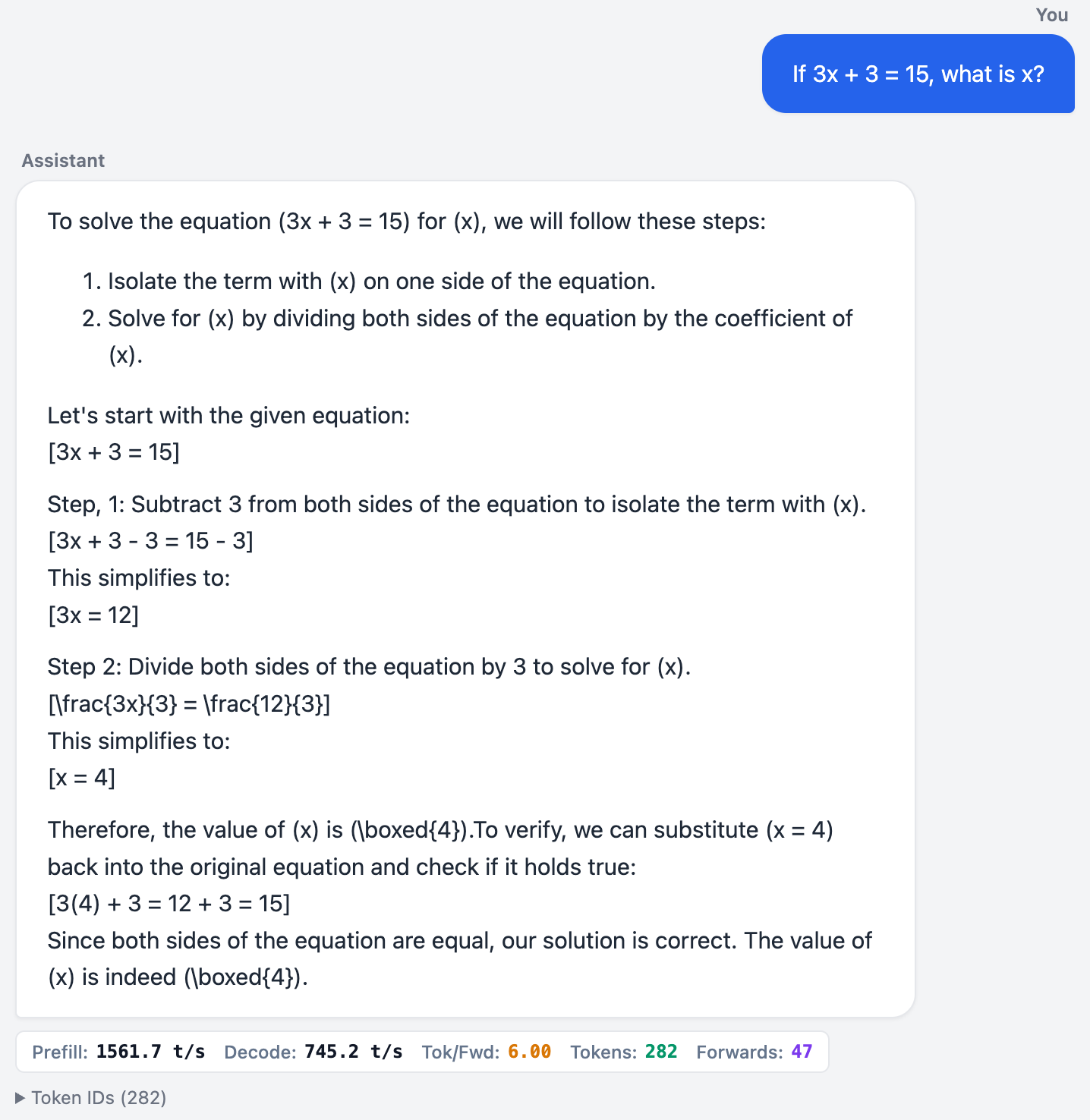}
    \caption{\textbf{Medium Entropy Case:} A mathematical reasoning task solving a linear equation. The structured nature of the step-by-step derivation allows for significant parallel decoding, resulting in \textbf{745.2 tokens/s}.}
    \label{fig:case_math}
\end{figure*}

\begin{figure*}[h]
    \centering
    \includegraphics[width=0.9\linewidth]{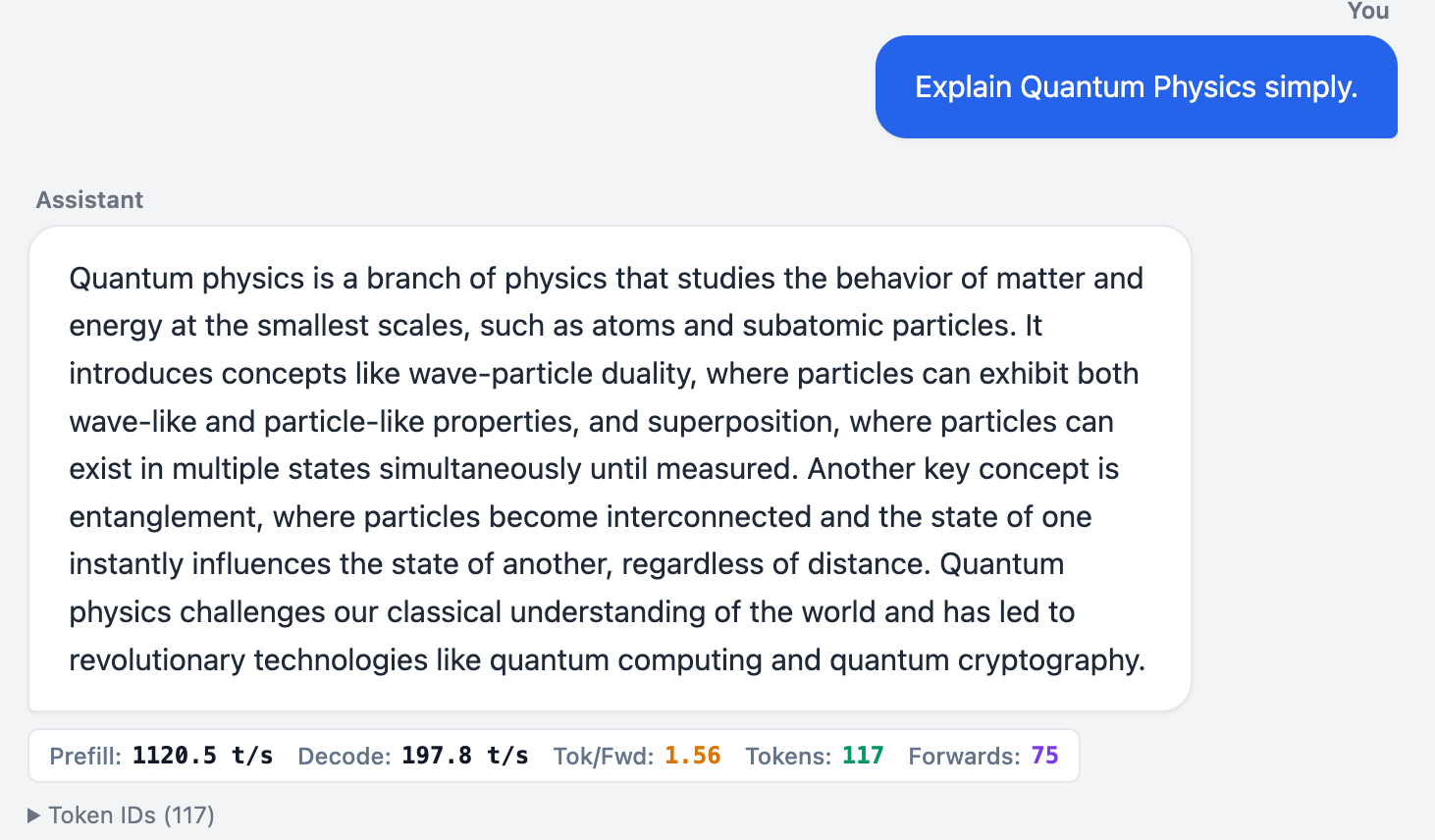}
    \caption{\textbf{High Entropy Case:} An open-ended knowledge explanation (Quantum Physics). The high semantic diversity and need for precise lexical selection reduce the effective parallel block size, resulting in a speed of \textbf{197.8 tokens/s}.}
    \label{fig:case_qa}
\end{figure*}

\end{CJK*}
\end{document}